\documentclass[manuscript,screen,review=false,nonacm]{acmart}
\AtBeginDocument{%
  \providecommand\BibTeX{{%
    \normalfont B\kern-0.5em{\scshape i\kern-0.25em b}\kern-0.8em\TeX}}}

\usepackage{blindtext}

\usepackage{xpatch}

\makeatletter
\xpatchcmd{\ps@firstpagestyle}{Manuscript submitted to ACM}{}{\typeout{First patch succeeded}}{\typeout{first patch failed}}
\xpatchcmd{\ps@standardpagestyle}{Manuscript submitted to ACM}{}{\typeout{Second patch succeeded}}{\typeout{Second patch failed}}    \@ACM@manuscriptfalse% Also in titlepage
\makeatother

\renewcommand\footnotetextcopyrightpermission[1]{} % removes footnote with conference info
\setcopyright{none}
\begin{document}

\title{Investigating the impact machine emotion classification uncertainty displays have on the human decision making process}

\author{Balaram Panda}
\email{bpan575@aucklanduni.ac.nz}
\orcid{0000-0001-5977-8392}
\affiliation{%
  \institution{The University of Auckland}
  \city{Auckland}
  \country{New Zealand}
  \postcode{1010}
}

\renewcommand{\shortauthors}{Balaram Panda}

\begin{abstract}

Many papers explore how well computers are able to examine emotions displayed by humans and use that data to perform different tasks. However, there have been very few papers which evaluate the computers ability to generate emotion classification information in an attempt to help the user make decisions or perform tasks. This is a crucial area to explore as it is paramount to the two way communication between humans and computers. This research conducted an experiment to investigate the impact of different uncertainty information displays of emotion classification on the human decision making process. Results show that displaying more uncertainty information can help users to be more confident when making decisions.

\end{abstract}



\maketitle

\section{Introduction}
% Henry

Human-computer interaction is a well-explored topic that has had many papers published about computers' abilities to classify human emotions. However, in order for a computer to effectively communicate with the humans that use it, the computer should be able to accurately identify the emotions that are being displayed. This is an affective computing problem that directly relates to the human-computer interaction (HCI) world. This is an incredibly important topic because accurate emotion recognition could lead to a significant improvement in the experience humans have with technology. There is also an extremely broad application domain that this could be applied to. For example, it could be used to help determine how to respond to someone who is in crisis, through the likes of a chatbot.

Computers are rapidly improving their ability to recognize emotion to make this a reality. It very often does this through training a model on "natural language processing". This is effectively, an approach to determine emotional keywords in sentences to "understand" how language is used. This has a great number of applications. Aside from being used to classify emotion, it has also been used in situations to help generate code, read street signs and even in video games.

However, one of the difficulties lies in conveying this information back to humans in a useful and meaningful way. This information is never completely accurate, so we also need to inform the user of the uncertainty associated with the information it is also gathering. We will be exploring how best we can display this uncertainty display to the user to help them make decisions.

\subsection{Relevant Literature}
%Yuwei

\begin{figure}[h]
  \centering
  \includegraphics[width=0.7\linewidth]{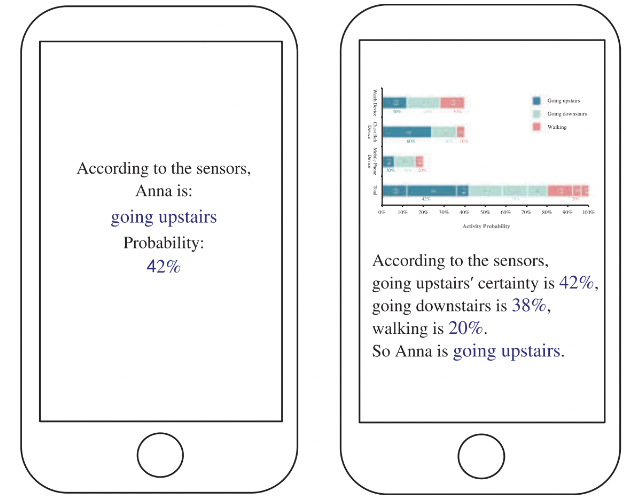}
  \caption{Two types of uncertainty visualisations shown to the group of participants.\cite{uncertain}}
    \label{fig:visualcomp}
\end{figure}

To supplement our motivations and explore the literature that exists relevant to our research question, there are three papers which can guide us for our project. Firstly, in \textit{Uncertainty Visualisation for Mobile and Wearable Devices Based Activity Recognition Systems}, the researchers investigated whether participants found percentage breakdowns of machine classification more intuitive than without. Figure \ref{fig:visualcomp} shows two designs: a simple visualisation showing just the probability, and the second screen displays more about how the classifier came to its conclusion with the percentage certainty at each stage of the activity. Participants found the horizontal bar chart with percentages much more intuitive, and many gave feedback about how that allowed them to build trust in the system \cite{uncertain}. 

\begin{figure}[h]
  \centering
  \includegraphics[width=0.5\linewidth]{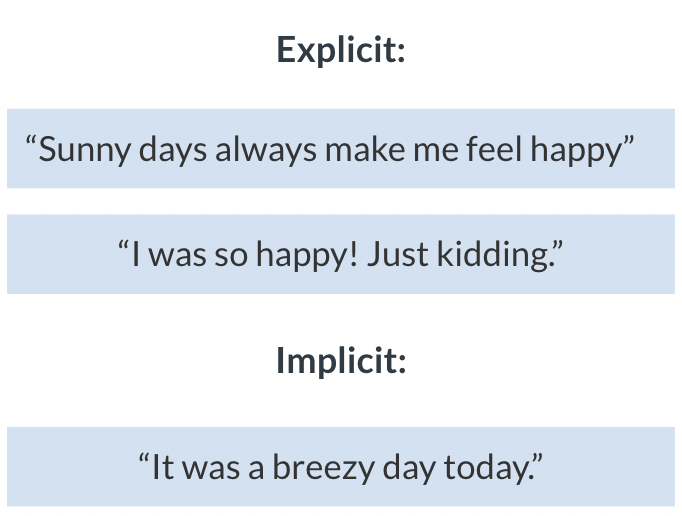}
  \caption{Explicit v.s. Implicit emotions.\cite{survey}\cite{fiction}}
    \label{fig:explicitimplicit}
\end{figure}

As our project is focused on emotion recognition in text, the second relevant paper we will mention is \textit{A Survey of state-of-the-art Approaches for Emotion Recognition in Text} which talks about the difficulties in classifying explicit and implicit emotions in text \cite{survey}. Explicit emotions are stated in the sentence directly, for example the word “happy” is used in the first sentence “sunny days always make me feel happy”. However, the difficulty for the classifier lies in needing to understand the entire phrase, such as in the second example in Figure \ref{fig:explicitimplicit}, the classifier must understand that “just kidding” negates the happiness expressed in the first part. Implicit emotions do not incorporate the emotion, and the classifier requires connotation and other training to improve accuracy. 
Key takeaways which informed how our machine learning classifier should behave and :
\begin{enumerate}
    \item Entire phrases must be analysed, not just the presence of a word
    \item Explicit emotions have different meanings in fiction and non-fiction texts: in fiction texts, emotions stay true to their meaning; however, in non-fiction texts, the emotion may just be stated and not relate to its meaning at all \cite{survey} \cite{fiction}. For example, in the phrase "The feeling of happiness is due to their chemicals in the human brain", it does not imply happiness.
    \item Implicit emotions require connotation training
    \item Feeding simple emotions into the AI text generator generates explicit phrases, whereas more complex emotions as keywords generates more implicit phrases. 
\end{enumerate}

\begin{figure}[h]
  \centering
  \includegraphics[width=0.5\linewidth]{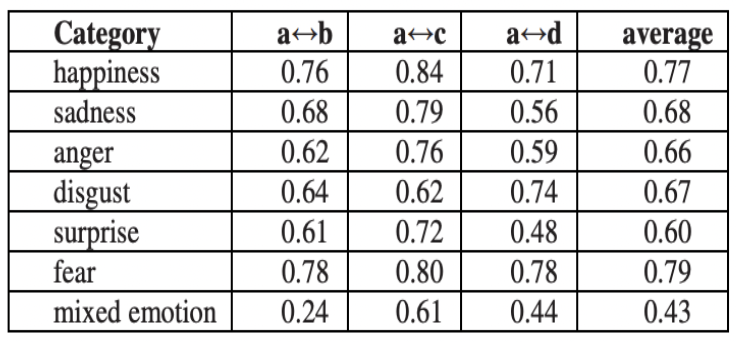}
  \caption{Comparison of subject A’s annotations with the other subjects. Values are how similar they are to the respective subject. \cite{identifyExpressions}}
    \label{fig:expressionResults}
\end{figure}

The paper \textit{Identifying Expressions of Emotions in Text} answers a similar question to ours \cite{identifyExpressions}. Their goal was to see how differently four judges annotate the same pieces of text and compare the similarity between them. Figure \ref{fig:expressionResults} shows how similar the subjects annotated certain emotions. It indicates that happiness and fear were the two emotions most collectively recognised and that surprise is the lowest. The experimental design was also novel in the addition of a seventh category - “mixed emotion” - which however, led a lot of responses to vary as it encouraged confusion.

\section{Research Question}
In our research, we wanted to further investigate the impact that showing machine uncertainty to a human user could have on their decision making process. This is to help understand how we can harness emotion recognition in computers to improve the fit between humans and computers.

As a result, our final research question is: "How does the display of machine uncertainty affect a human user's response and uncertainty toward the classification of emotion in written text?”

\section{Methods}
% Etienne

To address our research question, we performed an experiment with the goal of examining how machine uncertainty could affect human responses towards emotions presented in written text. To perform this experiment, we asked participants a series of questions about a range of stories generated with different emotions displayed in them to see if we found any differences between the group who had uncertainty data presented to them and those who did not. To generate the stories, we used GPT-3 \cite{gpt3}; GPT-3 is a natural language processor which has been trained on the largest data set for any natural language processor, this means it is able to interpret text and provide a reasonable response to text based prompts. We wrote a small script to call the API with a prompt with the structure "Tell me a <EMOTION> story about a <SUBJECT> with a <OBJECT>", such as "Tell me a happy story about an elf with a ball". using this structure, we were able to create many different stories about many characters. Initially, we directly fed these to the users. However, we found that GPT-3 sometimes did not complete the story leaving it with a half-completed sentence or would create a story about a sensitive topic that we did not want to expose the users to unnecessarily. As such, we filtered the stories down to 10, which were stored on Firebase \cite{firebase}, and could access to present to the users. This also had the added benefit of letting us explore trends across different stories. We used Hugging Face \cite{huggingface} to then evaluate the stories to test if the emotion presented was accurate to the prompt and to create probabilities of each story presenting different emotions. Hugging Face is a powerful classifier which is able to accurately detect the emotions displayed in text samples \cite{huggingface}. 

\begin{figure}[h]
  \centering
  \includegraphics[width=\linewidth]{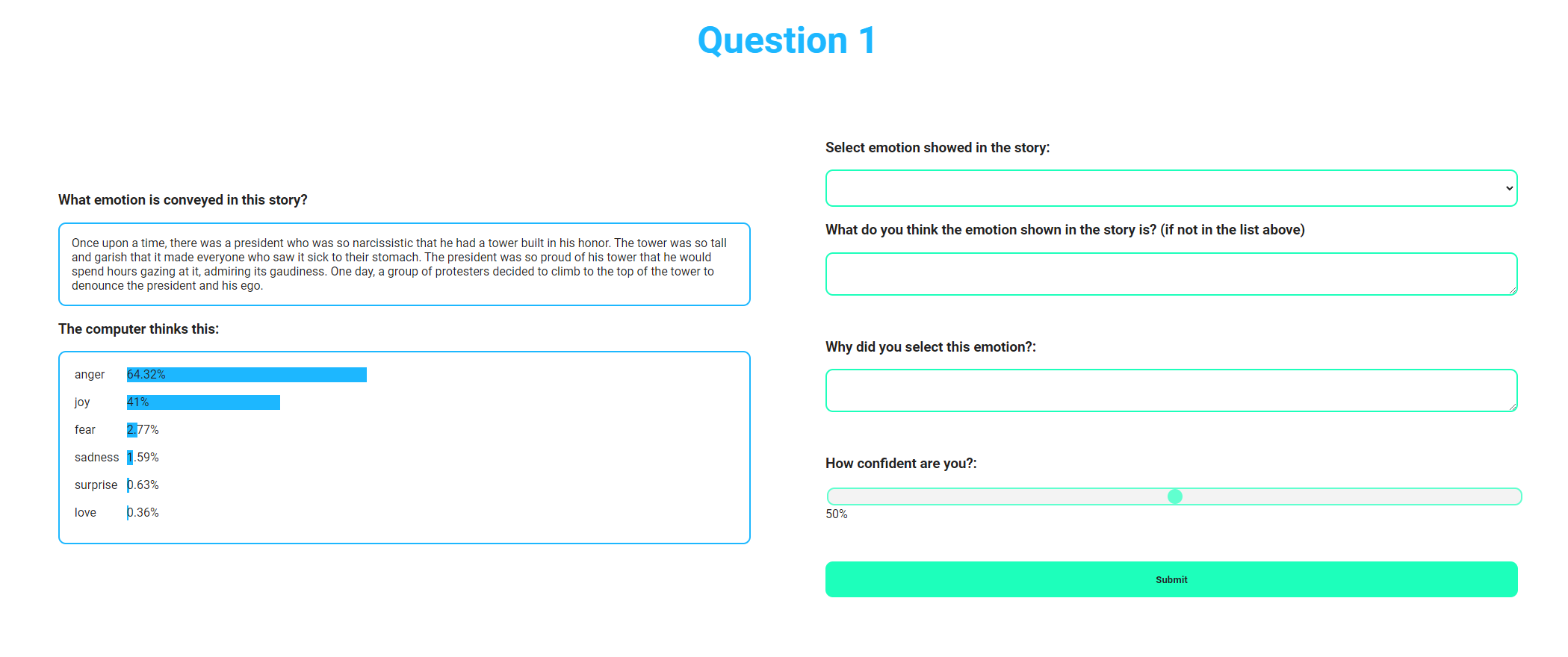}
  \caption{A display from the interface presented to the survey participants.}
    \label{fig:webapp}
\end{figure}

From this data, we created an open-source web app hosted at \url{https://github.com/etinaude/gpt3-hci-research} for users to answer our survey using Svelte \cite{svelte} and hosted on GitHub \cite{github}. Each user was asked for their unique identifier (a number between 0 and 100) and was presented with a consent form which they needed to agree to in order to continue. This consent form is shown in the appendix. A randomised control trials design was used. Once Users had completed this, they were split into two groups depending on their randomly allocated unique identifier, if the identifier was under 50, they were in group A (the control group); otherwise, they were in group B (the treatment group). Group A was presented with a simple prediction such as "The computer thinks this is: sad", while group B was provided with a breakdown of probabilities as shown in Figure \ref{fig:webapp}. The interface presented to the user was broken down into two sections; the first section was outlined in blue and displayed on the left; it shows computer-generated content, such as the story and the computer prediction. The second section contains the various input fields for the user to complete, this section was outlined in green. The second section had five elements in it, which are as follows:

\begin{enumerate}
  \item A drop-down menu of the emotion displayed in the story.
  \item A custom field for users to enter an emotion that they thought was displayed if the drop-down options did not accurately portray the emotion in the story.
  \item A text field for the user to explain why they picked the emotion they did.
  \item A confidence slider so we could explore how confidently the users were making these choices.
  \item A submit button once the user had completed the rest of the form.
\end{enumerate}

We surveyed 25 people of which 21 participants completed the entire survey, the remaining 4 data points were discarded. The participants ranged in age from 20 years old to 32 years old with a ratio of 6:4 male:female. Each participant answered 10 questions using our interface, as such, we were able to collect 210 data points. After the randomised allocation, 11 of these participants were in group A, and 10 were in group B.

\section{Results}
%Bala%
% How data being analysed
% Result comparison how responses varies 
%add figure

The experiment data was collected and stored in the Firebase cloud. The data was extracted into JSON file format and then processed using Python to transform into tables. Both quantitative and qualitative analyses were performed on the data and the results are as the following.

\begin{figure}[h]
  \centering
\includegraphics[width=\linewidth]{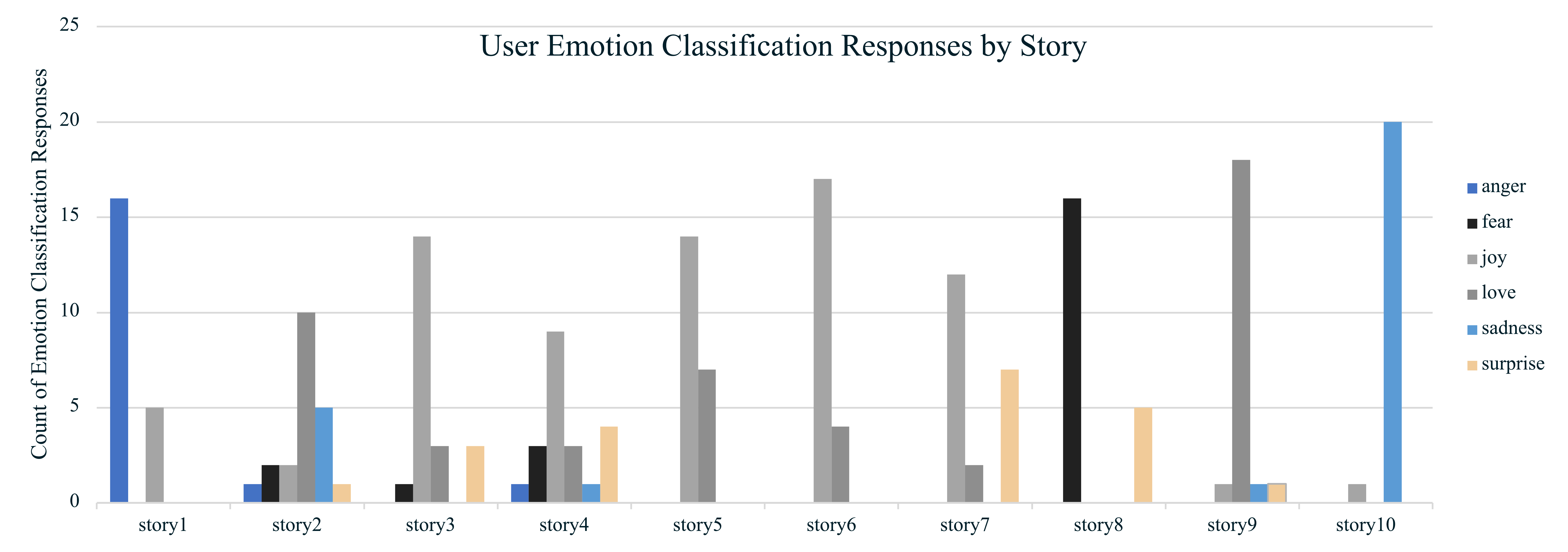}
  \caption{Various emotional responses on stories}
  \label{fig:emotion_responses}
  \Description{various emotional response on each story across group.}
\end{figure}

In Figure \ref{fig:emotion_responses}, the horizontal bar chart shows the emotion responses from participants, across various stories falling under different emotion categories. From this chart we can see the emotion responses to stories 2, 3, 4 and 9 varies across different emotion categories. Hence we can conclude that story 2, 3, 4 and 9 are have more uncertainty when classifying into a single, clear emotion category by the participants. 

\begin{figure}[h]
  \centering
  \includegraphics[width=0.8\linewidth]{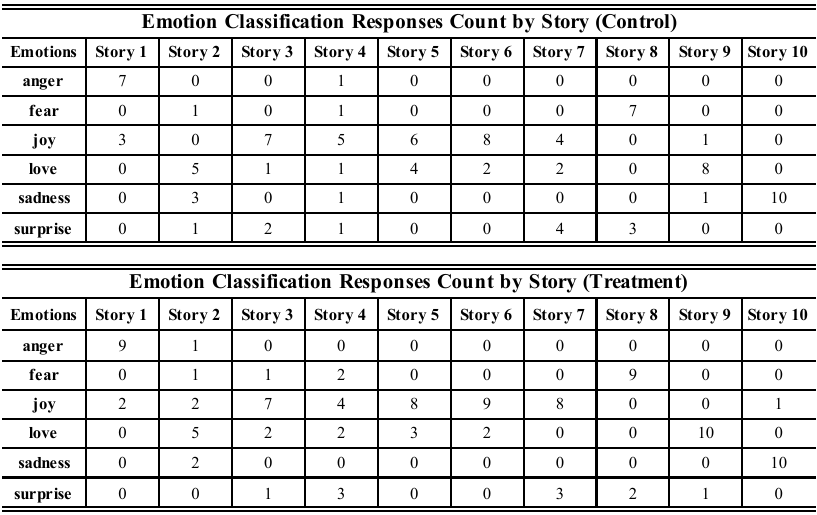}
  \caption{emotional responses by groups by stories}
  \label{fig:emotion_counts_by_grpup}
  \Description{various emotional response on each story from Treatment Group and Control Group}
\end{figure}

In Figure \ref{fig:emotion_counts_by_grpup} the table represents the response of each story from the Treatment Group and Control Group. From a close observation it can be found for some stories responses across the two groups vary. Following Figure \ref{fig:kappascore} helps to quantify that variation in emotion responses across two groups.   

%comparison of responses on each questions between two group
\begin{figure}[h]
  \centering
  \includegraphics[width=0.9\linewidth]{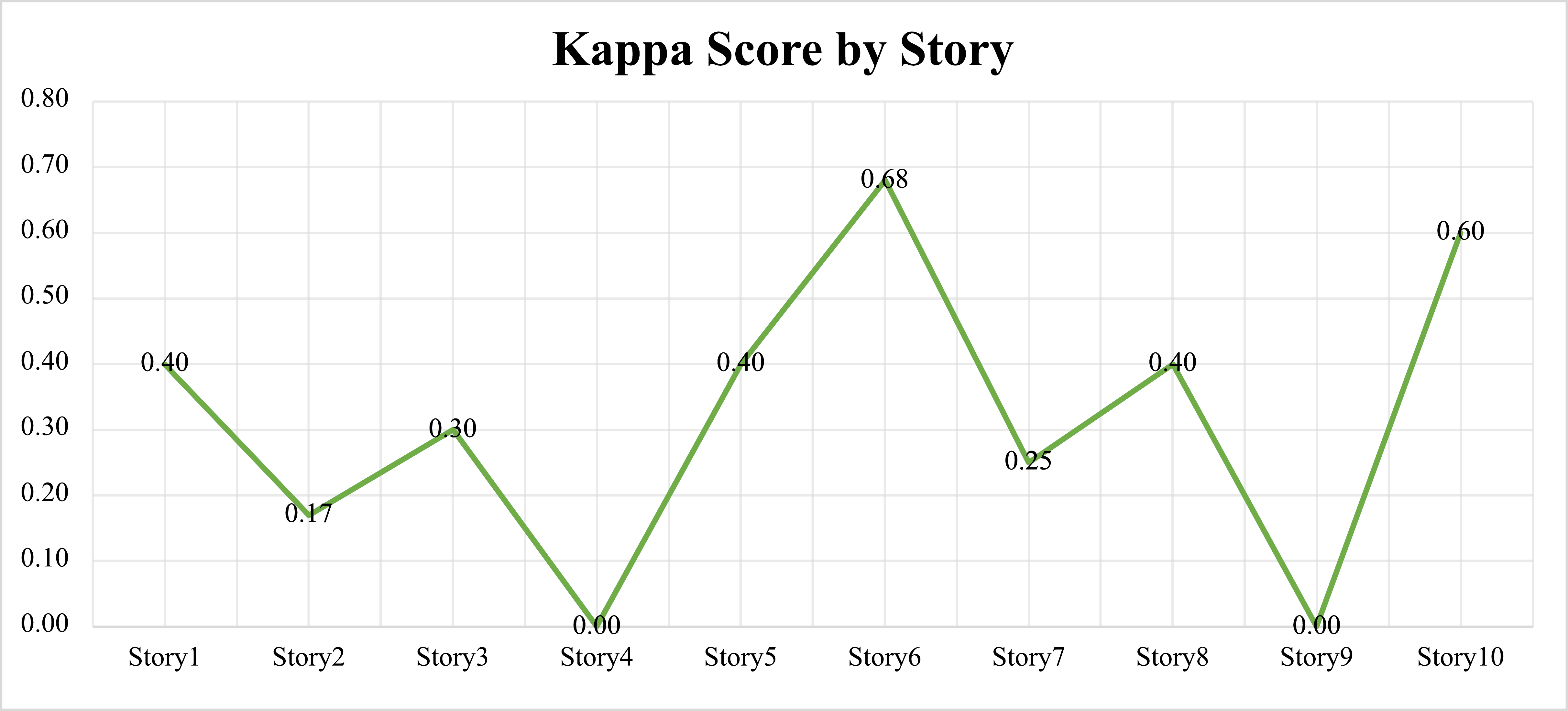}
  \caption{Kappa Score of Treatment Group and Control Group responses on each story}
  \label{fig:kappascore}
  \Description{Kappa Score calculated on the various responses from the participants of both groups(Treatment Group and Control Group) on each story.}
\end{figure}

\begin{figure}[h]
  \centering
  \includegraphics[width=0.4\linewidth]{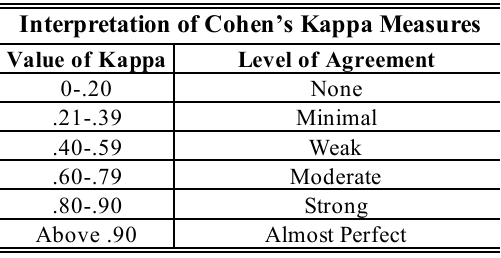}
  \caption{Interpretation of Kappa Statistics Measures \cite{mchugh2012interrater} }
  \Description{Interpretation Kappa based on the kappa range}
\end{figure}

Figure \ref{fig:kappascore} is a visualization of uncertainty which represents a comparison of responses across two groups on various stories using Kappa statistics applied on the data table mentioned in Figure \ref{fig:emotion_counts_by_grpup}. Cohen's kappa statistic \cite{mchugh2012interrater} measures inter-rater reliability (sometimes called inter-observer agreement). 
Kappa scores in Figure \ref{fig:kappascore} has been calculated comparing various emotional responses on each story then measuring the kappa of two groups. From the Kappa score we can see emotional responses vary from one group to another on the same story.
Based on Kappa score we can say most disagreement in emotion between two groups happens on stories 2, 4 and 9. 
Where as on stories 6 and 10, both groups agreed on their emotion responses up-to moderate level.
From Kappa score we can conclude that the story 2, 4 and 9 responses are more uncertain. This matches the uncertainty which was found from the "User Emotion Classification Responses by Story" Figure \ref{fig:emotion_responses}.

% Outcome: 1)  Uncertainty around responses on each question, quantification of uncertainty on those responses. 
% 2) Responses varies between groups and how to measure that
% 3) Link 1 and 2
% -- Bala's write up completed-- 

%Raina%
The Mann-Whitney U Two-tailed test on the self-reported confidence from the Control and Treatment groups was conducted to compare the user confidence in their emotion classification answers for each story. As shown in Table 1, results reveal that there are no statistically significant differences for most stories except story 7. However, there was a statistically significant difference found if data from all stories from each group is combined. In addition, comparing the confidence mean for two groups for story 7, the Treatment group has a much higher mean (7.55) than the Control group (5.20). Overall, the Treatment group has a higher confidence mean than the Control group most of the time. 
\begin{table}[h]
  \centering
  \includegraphics[width=0.75\linewidth]{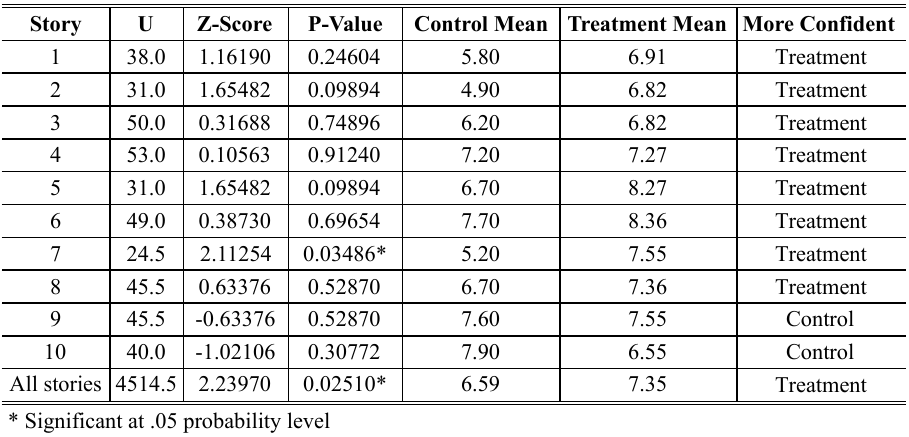}
  \caption{Mann-Whitney U Two-tailed Test of Self-reported Confidence and Means for Control and Treatment Groups}
\end{table}

\begin{figure}[h]
  \centering
  \includegraphics[width=\linewidth]{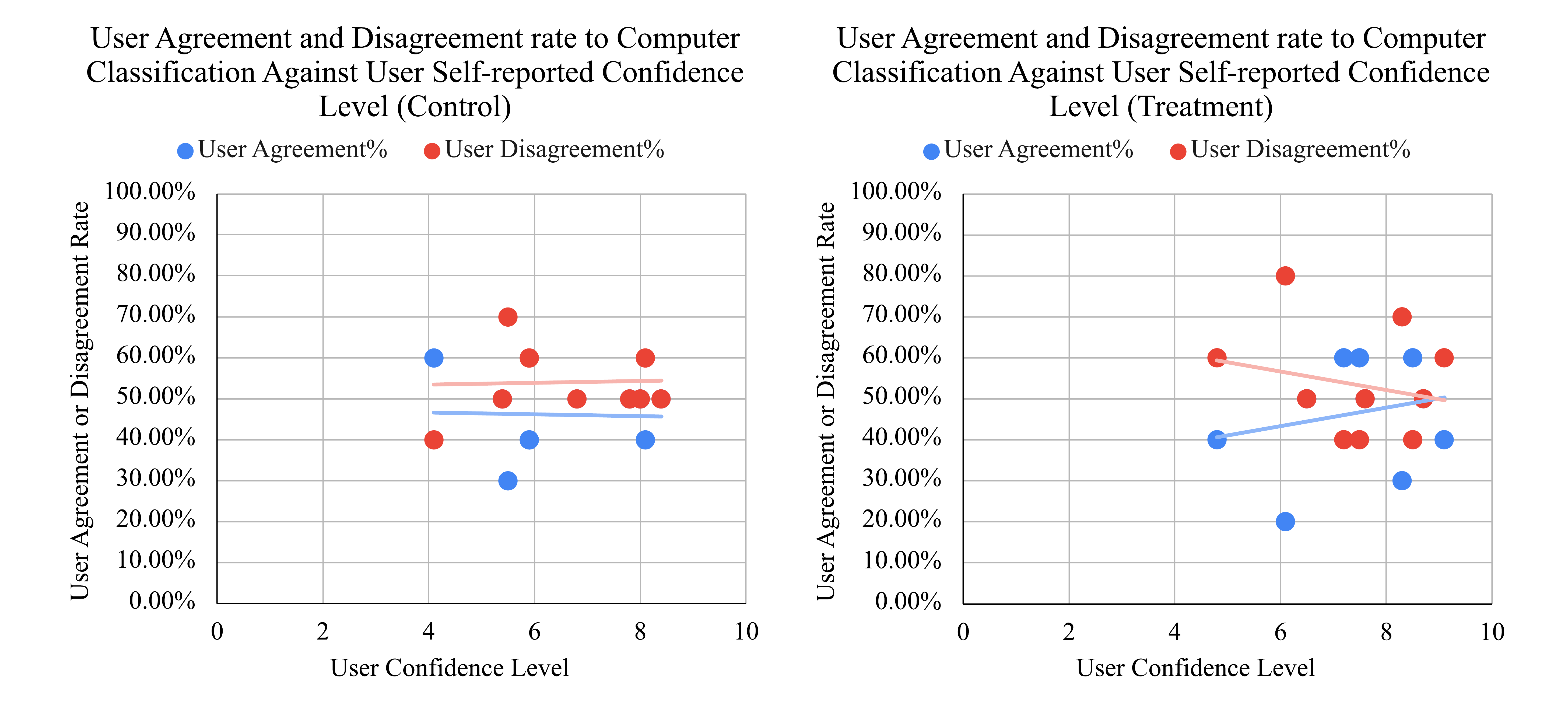}
  \caption{User Agreement and Disagreement rate to Computer Classification Against User Self-reported Confidence Level for the Control and Treatment Groups}
    \label{fig:scattergraph}
\end{figure}

The user confidence level is further evaluated against user's agreement and disagreement rates to computer classification results. During the experiment, users are able to see the computer's emotion classification(s). If the user's choice is the same as the computer's, we say they ``agree'' with the computer classification, and ``disagree'' otherwise. The results are represented in scatter graphs with trend lines as shown in the Figure \ref{fig:scattergraph}, the blue dots represent user's agreement rate and the red dots represent user's disagreement rate. User confidence level was reported on a scale of 0\% to 100\%, which translates to a scale of 0 to 10. The results show that there is a stronger correlation between the user confidence level and their agreement or disagreement to computer classifications in the Treatment group. In the Treatment group, the higher the confidence level the higher the agreement rate and the lower the disagreement rate. Indicating that users tend to be more confident when they agree with computer classifications in the Treatment group. On the other hand, there is little correlation found in the Control group. Moreover, comparing the two scatter graphs in \ref{fig:scattergraph}, the Treatment group has a wider spread scatter than the Control group.

The number of users based on their confidence level and agreement or disagreement with computer classification results was counted as shown in Table 2. The Treatment group was found to have more users have higher confidence than the Control group. Also, more users in the Treatment group agree or disagree with the computer classifications with higher confidence levels than the Control group. These results indicate that users in the Treatment group are more confident when they agreeing or disagreeing with computer classifications. In addition, most users in the study reported a higher than 50\% (or higher than 5 on a 1 to 10 scale) confidence level for most stories. In the 40 responses with lower confidence levels, most of the users are from the Control group.
\begin{table}[h]
  \centering
  \includegraphics[width=\linewidth]{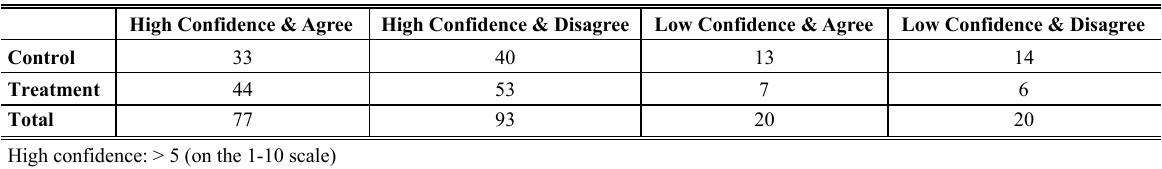}
  \caption{User Counts Based on Confidence Level and Agree or Disagree with Computer Classifications}
\end{table}

Table 3 shows a comparison between the Control and Treatment groups on the number of distinct emotion classifications extracted from the single choice and custom answer responses. If the number of emotion classifications in one group is more than the other group, we say the results in this group is more varied. For the single choice responses, there is not much differences in terms of variation between two groups. The two groups have an equal number of distinct emotion classification responses most of the time (five times) and when they are not equal the difference is also very small. However, the custom answer shows more variations, with the Control group has more variations than the Treatment group most of the times (seven times). In addition, the difference in number is more than two most of the times. Such results indicate that the Control group tend to have more variations in emotion classification responses than the Treatment group. In addition, another interesting finding was that users in the Treatment group tend to use more words to describe the emotions they felt in the stories, with an average of 35.79 characters from custom answer responses across all the stories, compared to 15.79 characters in the Control group. Although due to the limited number of responses there may be noises or errors in this result.
\begin{table}[h]
  \centering
  \includegraphics[width=0.8\linewidth]{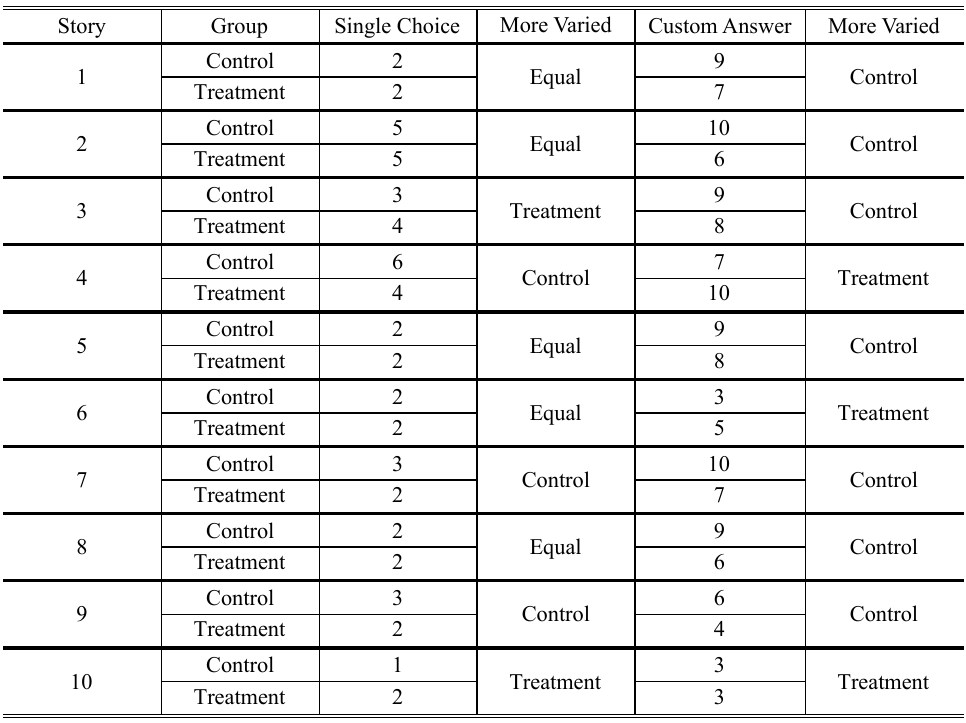}
  \caption{Distinct Emotion Classification Counts in Custom Answer and Single Choice Responses by Group (Emotion classification in one group is more varied if there are more distinct responses than the other group)}
\end{table}

% need to add another graph for thematic analysis

\begin{figure}[h]
  \centering
  \includegraphics[width=1.1\linewidth]{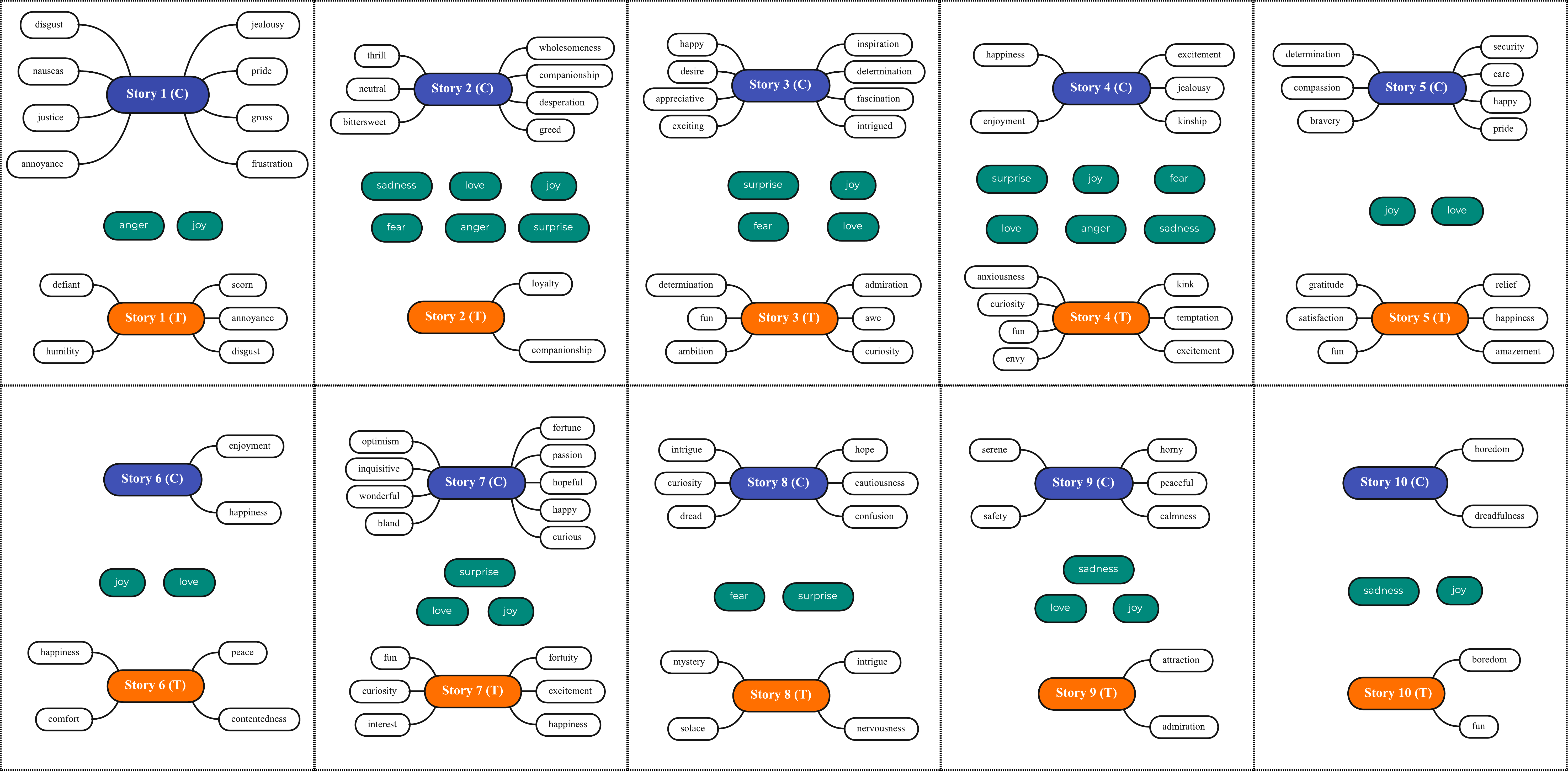}
  \caption{Distinct Emotion Classifications Extracted from Custom Answer Responses That Are Different From the Finite List of Emotions Provided (Blue: from the Control group; Orange: from the Treatment group; Green: single choice responses from two groups based on the finite list of emotions)}
  \label{fig:distinctEmotion}
\end{figure}

Figure \ref{fig:distinctEmotion} shows the distinct emotions classifications extracted from the custom answer responses that are different from the provided finite list of responses. For example, for story 1, users from two groups provided two different single choice responses "anger" and "joy". However, they provided in total 11 different words to describe the emotions they felt from reading the story. Such results indicate the limitation of computer emotion classification abilities.

\section{Discussion}
%Raina%
Some valuable results were obtained from the experiment which can be used to answer the research question at the start: "How does the display of machine uncertainty affect a human user’s response and uncertainty toward the classification of emotion in written text?". 

From the Kappa statistics measures shown in Figure \ref{fig:kappascore}, it can be seen that the inter-rater agreement measures between two groups are relatively high for story 6 and 10 but moderate or low for most of the stories. Such result could imply that user responses may vary based on the information displayed and thus two groups do not always agree on the same emotion classifications even though they were reading the same stories.

Moreover, the Mann-Whitney U Two-tailed test results shown in Table 1 did not report statistically significant differences of self-reported confidence level for most stories between two groups, but reported a statistically significant difference when all the data from each group is used. This could imply that we may not have enough data points to show statistical significant differences for each story as only around pieces of 11 data was used per group per story. Although no statistically significant differences were found for most stories, the Treatment group shows a higher mean for most stories which implies that the Treatment group is more confident than the Control group when classifying emotions for most stories. However, since there is only one confidence bar in the application and the instructions did not specify if this confidence bar is for the single choice or custom answer response, the results may have some noise or errors.

In addition, the correlation between the user confidence level and their agreement and disagreement rate to computer classifications was found in the Treatment group but not in the Control group. This result implies that the display of uncertainty information indeed have impact on user's decision making. Combining this finding with Table 2, which shows that more users in the Treatment group are more confident when agreeing and disagreeing with computer classifications, we could say that seeing more uncertainty information can help users to be more confident when making decisions. It was surprising to see that more uncertainty information can increase people's confidence, but such results can be supported by previous studies by Peterson and Pitz \cite{peterson1988confidence} that as the amount of information provided (uncertainty) increase, people's confidence in making predictions. 

Furthermore, from the results of distinct emotion classifications from two groups in Table 3 we can see that users in the Control group tend to provide more variations in terms of different words to describe the emotions they felt in the stories. This result could imply that due to the limited information displayed to the Control group, users are less certain on their classification results, thus their custom answers tend to be more uncertain. However, from the finding that the users in the Treatment group tend to use more words to describe the emotions in the custom answer responses, we can also see that when more uncertainty information is provided, users tend to be less concise in their answers as they have more information to process. 

Looking at the distinct emotion classification extracted from the custom responses that are different from the provided finite list of emotions in Figure \ref{fig:distinctEmotion}, there are some interesting results. The huge difference in the number of description words provided in the single choice and custom answer responses reflect that the common emotion classifications from the computer are still not precise enough to capture all the complex emotions that humans normally capture. However, we also can see that the words used are not always emotions, users often confuse words that describe emotions with those describe actions. For example, users used "loyalty", "companionship" to describe the complex emotions in story 2, but these words are not precisely emotions. Thus, there could be an improvement made to the instructions of the experiment.

Overall, from the results above we can answer the research question that by showing more uncertainty information to users, they tend to be more confident in making decisions about emotion classifications. 

\subsection{Threats to Validity}
There were several factors that may have skewed our results and have been a threat to the validity of our results.
\newline
The first was our sample size. In the experiments that were run, there were 10 participants in each of the control and the treatment groups. In order to increase the confidence of our results, we would need to increase the sample size. The increase in data points from a larger number of participants would allow us to more easily separate signal from noise in coming to conclusions about the meaning of our results.
\newline
The next threat to validity was the type of participants that took part in the experiment. As we reached out to classmates, friends and family this could have heavily skewed the type of participants that we collected results from. They tended to be from the ages of 20-24, university educated, with there being more males than females. This lines up with studies done on WEIRD biases, which can greatly affect the results of studies \cite{weird}. Our results are therefore actually more focused on a smaller population of people. Having a wider spread in the backgrounds of the people who take part in the experiment may reveal new insights or different results to what currently exists and increase external population validity.
\newline
Another factor within the participants to consider that was not overall assessed in our existing participants was their level of English comprehension. The act of participants reading, understanding and classifying the emotion of a machine generated excerpt is understandably nuanced and not easily achieved by those who have a weak grasp of the English language. A lack of understanding could have greatly affected their classifications and results. This was a quality of the participants that should have been tracked and potentially controlled.

\section{Conclusion}
We began by posing the research question of understanding how the display of machine uncertainty affected a human user's response and uncertainty toward the classification of emotion in written text. From our results, we generated two main conclusions. The first is that participants in the Treatment group, who were shown the machine emotion classification uncertainty information display tended to be more confident in their responses than the Control group. The second is that while the Treatment group was more confident when making emotion classifications, they also disagreed with machine classifications more than the Control group did. Overall, we can conclude that the machine emotion classification uncertainty displays indeed have an impact on the human decision making process, the more uncertainty information is displayed, the more certain humans are when making decisions, even when they are disagreeing with what they were told by the machine, and also the less varied their answers are. In addition, we have found that the machine emotion classification ability is still not accurate enough to capture complex emotions that humans could capture easily, which opened up opportunities for future research.

The research done was an insightful look at affective computing in the HCI field and a great first step in understanding how the humans can more effectively interact and interface with technology.

\subsection{Future Work}
As the research into uncertainty and emotion classification has only been loosely explored, there are several areas which we could work and build off of. 

The first would be looking at changing the type of uncertainty display that is shown to the participants and building a strong understanding of how different types of displays and the amount of information revealed to the participants affected results. For example, changing the existing model to another prominent uncertainty display model and understanding how this affects participant response and uncertainty would be fascinating and an interesting aspect to build upon and study.

A change that could be explored is the generation of the written texts that are classified by participants. In the experiment they were done by GPT-3, but exploring human generated written texts may prove to show interesting results in how we understand the texts written by other humans rather than machines.

Another avenue that could be explored would be into other methods of emotion classification. In our study, we have only focused on written emotion classification, but understanding other methods such as vocal, facial-visual or physiological modes of emotion classification and how they affect participant classification and uncertainty would be equally as valuable and interesting.

\section{Acknowledgments}

We would like to acknowledge Dr Danielle Lottridge and Dr Gerald Weber for their continued support and helping us understand the complexities of human-computer interaction during the writing of this paper.

\bibliographystyle{ACM-Reference-Format}
\bibliography{references}

\appendix

\section{Consent form template}

THIS FORM WILL BE HELD FOR A PERIOD OF 2 MONTHS

Project title: Visualising and supporting uncertainty in emotion

recognition displays

%Name of Supervisor: Danielle Lottridge, Gerald Webber

%Name of Student Researchers: Etienne Naude, Henry Gann, Balaram Panda, Lance Zhang

I have read the Participant Information Sheet, have understood the nature of the research and why I have been selected. I have had the opportunity to ask questions and have had them answered to my satisfaction.

• I agree to take part in this research.

• I understand that I am free to withdraw my participation at any time, and to withdraw any data traceable.

• I wish / do not wish to receive the summary of findings.

• I agree to not disclose anything discussed in the focus group.

\section{PIS}

Thank you for agreeing to take part in our HCI experiment. Please ensure that you take part in this study completely alone. It will take approximately 10 minutes to complete and you will be asked to classify 10 short stories.

You will firstly be asked to enter your user ID - this should have been provided to you before the experiment. As your proceed from there you will be presented a story generated by a computer. This same story will also be inputted and classified by an AI and the emotion(s) it thinks is being displayed most will be shown to you.

Following this you will select from a finite list of emotions the emotion you think best matches what was conveyed in the story above.

In order for us to build a better understanding, we ask you to write a short answer reasoning why you selected the emotion.

Finally, select on the slider the confidence level of your answer. 0\% (left) being not confident at all and 100\% (right) being very confident.

\section{Questionnaire}

\subsection{Stories}

\subsubsection{Story 1}

Once upon a time, there was a president who was so narcissistic that he had a tower built in his honor. The tower was so tall and garish that it made everyone who saw it sick to their stomach. The president was so proud of his tower that he would spend hours gazing at it, admiring its gaudiness. One day, a group of protesters decided to climb to the top of the tower to denounce the president and his ego.

\subsubsection{Story 2}
Once upon a time, there was a thief who had a dog. The dog was his only friend and companion. The thief loved his dog dearly and would do anything for him. One day, the thief decided to rob a wealthy home. He knew that the homeowners kept a lot of cash in their safe. He planned to break in while they were away and take the money. The thief’s dog accompanied him on the job.

\subsubsection{Story 3}
Fred was always admiring the way that other people could use a hammer so effectively. He would watch them intently as they would drive nails into a piece of wood or metal, and he would always be amazed at how they made it look so easy. He knew that it couldn't be easy, but he was still fascinated by it. One day, he decided that he was going to learn how to use a hammer himself.

\subsubsection{Story 4}
Fred was a man who loved to play with balls. He would spend hours upon hours playing with them, bouncing them, throwing them, catching them. He was a master of the ball. One day, while Fred was playing with his ball, he heard a voice call out to him. "Hey Fred! Come play with us!" The voice came from a group of people who were playing with a different kind of ball. It was a bigger ball.

\subsubsection{Story 5}
There once was a wizard who had a hammer. He was a great wizard and used his hammer to help others. He always had a helping thand when someone needed it and was always there to lend a helping hand. One day, a group of bandits attacked a village and the wizard used his hammer to help defend the village. The villagers were very grateful and the wizard was hailed as a hero.

\subsubsection{Story 6}
Mary was a kind and gentle lady who loved nature. She would often go for walks in the park and admire the trees. One day, she was walking in the park and she saw a beautiful tree. She asked the park ranger if she could have the tree planted in her honor. The park ranger said yes, and Mary was so happy. The tree was planted and Mary would visit it often. She would sit under the tree and read or just enjoy the peace

\subsubsection{Story 7}
Once upon a time there was an archer who had a ball of yarn. She would often find herself daydreaming about all the different ways she could use it. She could make a bow string, a quiver, or even a blanket. The possibilities seemed endless. One day, she decided to take her ball of yarn to the local market to see what she could trade it for. She ended up meeting a woman who was looking for a new bow string.

\subsubsection{Story 8}
The pixie was floating along on her balloon, high above the treetops, when she spotted a colorful butterfly flitting about below her. She followed the butterfly as it flitted from flower to flower, until it suddenly disappeared into a dark cave. The pixie floated down to the entrance of the cave and peered inside. It was very dark and she couldn't see anything. She hesitated for a moment, but then she heard the butterfly's voice calling to her from inside.

\subsubsection{Story 9}
Fred was a young boy when he first built his treehouse. It was his special place, where he would go to be alone with his thoughts. As he grew older, he continued to use his treehouse as a place to escape the hustle and bustle of the outside world. One day, when Fred was in his treehouse, he saw a young woman walking by. Her name was Mary, and she was the most beautiful person Fred had ever seen.

\subsubsection{Story 10}
The president was having a very boring day. He woke up, ate breakfast, and then went to his desk to work. He signed a few documents, made a few phone calls, and then got up to stretch his legs. He walked around the Oval Office a few times, and then sat back down at his desk. He started doodling on a piece of paper, and then started daydreaming. He thought about what it would be like to go outside and play.

\end{document}